\documentclass[runningheads]{llncs}
\usepackage{graphicx}
\usepackage{amsmath,amssymb} 
\usepackage{color}
\usepackage[width=122mm,left=12mm,paperwidth=146mm,height=193mm,top=12mm,paperheight=217mm]{geometry}

\usepackage{supertabular}
\usepackage{dcolumn}
\usepackage{multirow}
\usepackage{rotating}
\usepackage{subfigure}

\usepackage{mathtools}
\usepackage{booktabs} 
\usepackage{array}
\usepackage[table]{xcolor}
\usepackage{gensymb} 
\usepackage{float}
\usepackage{soul}
\usepackage{layouts}
\usepackage{floatpag} 
\usepackage{nicefrac}
\usepackage{dirtytalk}
\usepackage{enumitem} 

\newcommand{\abs}[1]{\left\vert#1\right\vert} 
\newcommand{\myurl}[1]{{\footnotesize\url{#1}}}
\newcommand{\colorcell}{\cellcolor{gray!20}}
\newcolumntype{C}[1]{>{\centering\let\newline\\\arraybackslash\hspace{0pt}}m{#1}}

\begin{document}
\pagestyle{headings}
\mainmatter

\title{How useful is photo-realistic rendering for visual learning?} 

\titlerunning{How useful is photo-realistic rendering for visual learning?}

\authorrunning{Yair Movshovitz-Attias, Takeo Kanade, Yaser Sheikh}

\author{Yair Movshovitz-Attias, Takeo Kanade, Yaser Sheikh}



\institute{Computer Science Department \& Robotics Institute,\\
    Carnegie Mellon University\\
    \email{ \{yair,Takeo.Kanade,yaser\}@cs.cmu.edu}
}

\maketitle

\begin{abstract}
Data seems cheap to get, and in many ways it is, but the process
of creating a high quality labeled dataset from a mass of data is time-consuming
and expensive.

With the advent of rich 3D repositories, photo-realistic rendering systems offer
the opportunity to provide nearly limitless data. Yet, their primary value for
visual learning may be the quality of the data they can provide rather than the
quantity. Rendering engines offer the promise of perfect labels in addition to
the data: what the precise camera pose is; what the precise lighting location,
temperature, and distribution is; what the geometry of the object is.

In this work we focus on semi-automating dataset creation
through use of synthetic data and apply this method to an
important task -- object viewpoint estimation. Using
state-of-the-art rendering software we generate a large labeled dataset of cars
rendered densely in viewpoint space.
We investigate the effect of rendering parameters on estimation performance
and show realism is important.
We show that generalizing from synthetic data is not harder than the domain
adaptation required between two real-image datasets and that combining synthetic
images with a small amount of real data improves estimation accuracy.
\end{abstract}

\section{Introduction}
\label{sec:introduction}
The computer vision community has been building datasets for decades, and as
long as we have been building them, we have been fighting their
biases. From the early days of COIL-100~\cite{nene1996columbia} the Corel
Stock Photos and 15 Scenes datasets~\cite{oliva2001modeling} up to and including
newer datasets such as PASCAL VOC~\cite{pascal-voc-2011} and
Imagenet~\cite{imagenet_cvpr09}, we have experienced bias:
every sample of the world is biased in some way -- viewpoint, lighting, etc.
Our task has been to build algorithms that perform well on
these datasets. In effect, we have ``hacked'' each new dataset - exploring it, identifying weaknesses, and in sometimes
flawlessly fitting to it.

\begin{figure}[t!]
  \centering
    \includegraphics[width=0.65\textwidth]{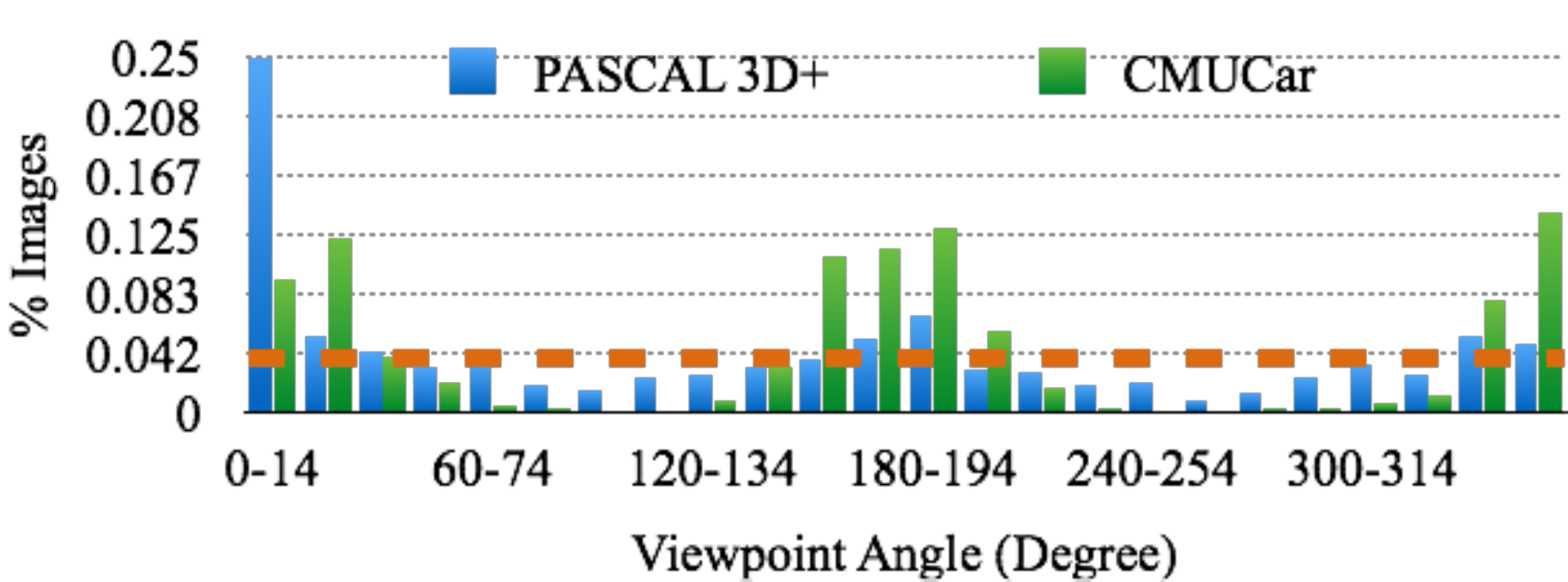}
    \caption{Photographers tend to capture
    objects in \emph{canonical} viewpoints.
    When real images are used as training data these
    viewpoints are oversampled. Here we show viewpoint distributions for two real
    image datasets. Note the oversampling of certain views, e.g. $0\degree$, $180\degree$.
    In comparison, an advantage of a
    synthetic dataset is that it is created to
    specification. A natural distribution to create is
    uniform (dashed line).}
    \label{FIG:angle_freq_compare}
\end{figure}

For an in depth analysis of the evolution of datasets (and an enjoyable read) we
refer the reader to~\cite{torralba2011unbiased}. In short, there
are two main ways in which our community has addressed bias: making new
datasets, and building bigger ones. By making new datasets we continuously
get new samples of the visual world, and make sure our techniques handle more of
its variability. By making our datasets larger we make it harder
to over-fit to the dataset's individual idiosyncrasies.

This approach has been remarkably successful. It requires, however, a great
amount of effort to generate new datasets and label them with ground truth
annotations. Even when great care has been taken to minimize the sampling bias
it has a way of creeping in. As an example, for the task of object viewpoint
estimation, we can observe clear bias in the distribution viewpoint angles when
exploring real image datasets. Figure~\ref{FIG:angle_freq_compare} shows the
distribution of azimuth angles for the training sets of the car class of PASCAL
VOC, and the CMUCar dataset. There is clear oversampling of some angles, mainly
around $0\degree$ and $180\degree$.

In this work we explore the benefits of synthetically generated data for
viewpoint estimation. 3D viewpoint estimation is an ideal task for the use of
renders, as it requires high of accuracy in labeling. We utilize a large
database of accurate, highly detailed, 3D models to create a large number of
synthetic images. To diversify the generated data we vary many of the rendering
parameters. We use the generated dataset to train a deep convolutional network
using a loss function that is optimized for viewpoint estimation. Our
experiments show that models trained on rendered data are as accurate as those
trained on real images. We further show that synthetic data can be also be used
successfully during validation, opening up opportunities for large scale
evaluation of trained models.

\begin{figure}[t!]
  \centering
    \includegraphics[width=0.83\linewidth]{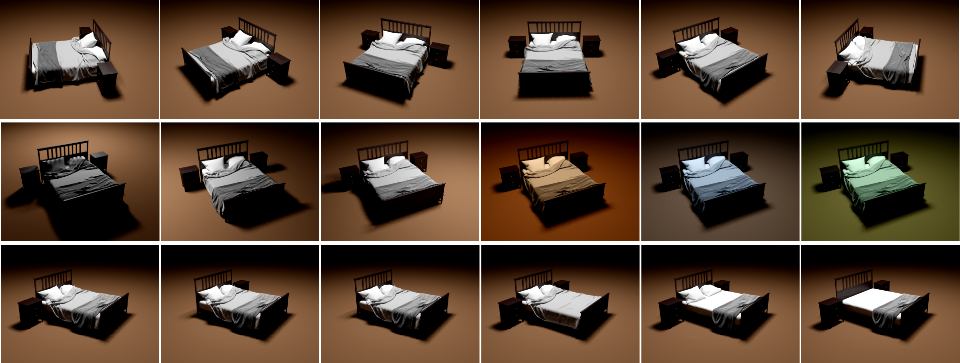}
    \caption{Generating synthetic data allows us
    to control image properties, for example, by: sampling
    viewpoint angles (top row), sampling lighting conditions (middle), and sampling occlusions
    by controlling the placement of night stands and
    linens (bottom).
    }
    \label{FIG:sofa_angles}
\end{figure}

With rendered data, we control for viewpoint bias, and can create a
uniform distribution. We can also adequately sample lighting conditions and
occlusions by other objects. Figure~\ref{FIG:sofa_angles} shows renders
created for a single object, an IKEA bed. Note how we can sample the different
angles, lighting conditions, and occlusions. We will, of course, have other
types of bias, and this a combined approach -- augment real image
datasets with rendered data. We explore this idea below.

We assert that a factor limiting the success of many computer vision
algorithms is the \emph{scarcity} of labeled data and the \emph{precision} of
the provided labels.
For viewpoint estimation in particular the space of possible angles is
immense, and collecting enough samples of every angle is hard. Furthermore,
accurately labeling each image with the ground truth angle proves difficult for
human annotators. As a result, most current 3D viewpoint datasets resort to one
of two methods for labeling data: (1) Provide coarse viewpoint information in
the form of viewpoint classes (usually 8 to 16). (2) Use a two step process of
labeling. First, ask annotators to locate about a dozen keypoints on the object
(e.g., front-left most point on a car's bumper), then manually locate those same
points in 3D space on a preselected model. Finally, perform PnP
optimization~\cite{Epnp09} to learn a projection matrix from 3D points to 2D
image coordinates, from which angle labels are calculated. Both methods are
unsatisfying. For many downstream applications a coarse pose classification is
not enough, and the complex point correspondence based process
expensive to crowdsource. By generating synthetic images one can
create large scale datasets, with desired label granularity level.

\section{Related Work}
\label{sec:related_work}
The price of computational power and storage has decreased dramatically over the
last decade. This decrease ushered in a new era in computer vision, one that
makes use of highly distributed inference
methods~\cite{dean_2012,szegedy_going_2014} and massive amounts of labeled
data~\cite{ILSVRCarxiv14,goodfellow_multi-digit_2013,movshovitzattias:2015:CVPR}.
This shift was
accompanied by a need for efficient ways to quickly and accurately label these
large datasets. Older and smaller datasets were normally collected and
labeled by researchers themselves. This ensured high quality labels but was not
scalable. As computer
vision entered the age of ``Big Data'' researchers began looking for better
ways to annotate large datasets.

Online labor markets such as Amazon's Mechanical Turk have been
used in the computer vision community to crowdsource simple tasks such as image
level labeling, or bounding box annotation
\cite{ILSVRCarxiv14,law_human_2011,von_ahn_labeling_2004}.
However, labor markets often lack expert
knowledge, making some classes of tasks impossible to complete. Experts are rare
in a population and when are not properly identified, their answers will be
ignored when not consistent with other workers -- exactly on the instances where
their knowledge is most crucial\cite{heimerl_communitysourcing:_2012}.

The issues detailed above make for a compelling argument for automating the
collection and labeling of datasets. The large increase in availability of 3D
CAD models, and the drop in the costs of obtaining them present an appealing
avenue of exploration: Rendered images can be tailored for many computer vision
applications. Sun and Saenko~\cite{sun2014virtual} used rendered images as a
basis for creating object detectors, followed by an adaption approach based on
decorrelated features. Stark et al.~\cite{stark2010back} used 3D models of cars
as a source for labeled data. They learned spatial part layouts which were used
for detection. However, their approach requires manual labeling of semantic part
locations and it is not clear how easy this can scale up to the large number of
objects now addresses in most systems. A set of highly detailed renders was used
to train ensembles of exemplar detectors for vehicle viewpoint estimation
in~\cite{movshovitzattias:2014:BMVC}. Their approach required no manual labeling
but the joint discriminative optimization of the ensembles has a large
computational footprint and will be hard to scale as well.
Pepik et al.~\cite{DBLP:journals/corr/PepikBRS15} showed that deep networks are
not invariant to certain appearance changes, and use rendered data to augment the training data, and in~\cite{hattori2015learning,vazquez2014virtual} rendered data is used to train pedestrian detectors.

Here, we show that detailed renders from a large set of high quality 3D
models can be a key part of scaling up labeled data set curation. This
was unfeasible just 10 years ago due to computational costs, but a
single GPU today has three
orders of magnitude more compute power than the server farm used by
Pixar for their 1995 movie Toy
Story~\cite{fatahalian_enolving_2011}. The time is ripe for
re-examining synthetic image generation as a tool for computer vision.
Perhaps most similar to our work is~\cite{su2015render} in which
rendered images from a set of 3D models were employed for the task of
viewpoint estimation. However, while they focus on creating an end-to-end
system, our goal is to systematically examine the benefits of rendered data.

\section{Data Generation Process}
\label{sec:data_generation}
To highlight the benefits of synthetic data we opt to focus on the car object
class. We use a database of 91 highly detailed 3D CAD models obtained from
doschdesign.com and turbosquid.com

For each model we
perform the following procedure. We define a sphere of radius $R$ centered at
the model. We create virtual cameras evenly spaced on the sphere in one degree
increments over rings at 5 elevations: $-5\degree, 0\degree, 10\degree,
20\degree, 30\degree$. Each camera is used to create a render of one viewpoint
of the object. We explore the following rendering parameters:

\noindent \textbf{Lighting position}
We uniformly sample the location of a directed light on
a sphere, with elevation in $[10\degree, 80\degree]$.

\noindent \textbf{Light Intensity} We uniformly sample the Luminous power (total emitted
visible light power measured in lumens) betwen 1,400 and 10,000. A typical 100W
incandescent light bulb emits about 1500 lumens of light, and normal day light
is between 5,000 and 10,000 lumens. The amount of power needed by the light
source also depends on the size of the object, and the distance of the light
source from it. As models were built at varying scales, the sampling of this
parameter might require some adjustments between models.

\begin{figure}[t!]
  \centering
  \includegraphics[width=0.8\linewidth]{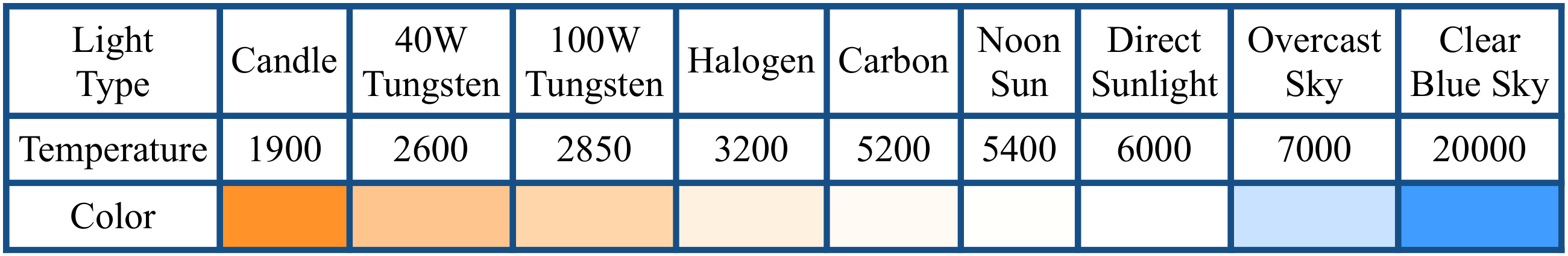}
  \caption{Set of light source temperatures used in rendering process.
  We create each render with a randomly selected temperature.
  }
  \label{FIG:light_temperature}
\end{figure}

\noindent \textbf{Light Temperature} We randomly pick one of $K=9$ light temperature
profiles. Each profile is designed to mimic a real world light scenario, such as
midday sun, tungsten light bulb, overcast sky, halogen light, etc (see Figure~\ref{FIG:light_temperature}).

\noindent \textbf{Camera F-stop} We sample the camera aperture F-stop uniformly between 2.7
and 8.3. This parameter controls both the amount of light entering the camera,
and the depth of field in which the camera retains focus.

\noindent \textbf{Camera Shutter Speed} We uniformly sample shutter speeds between
$\nicefrac{1}{25}$ and $\nicefrac{1}{200}$ of a second. This controls
the amount of light entering the camera.

\noindent \textbf{Lens Vignetting} This parameter simulates the optical vignetting effect of
real world cameras. Vignetting is a reduction of image brightness and saturation
at the periphery compared to the image
center. For 25\% of the
images we add vignetting with a constant radius.

\noindent \textbf{Background} Renders are layered with a natural image background
patches that are randomly selected from PASCAL training images not from
the ``Car'' class.

For rendering the images we use 3DS MAX, with the production quality
VRAY rendering plug-in~\cite{vray}.
There has been considerable evidence that data augmentation methods can
contribute to the accuracy of trained deep
networks~\cite{wu_deep_2015_deep_image}. Therefore, we augment our
rendered images by creating new images, as follows:

\noindent \textbf{Compression Effects} Most of the images that the trained
classifier will get to observe during test time are JPEG compressed images. Our
renders are much cleaner, and are saved in lossless PNG format. In most cases
JPEG compression is not destructive enough to be visually noticeable, but it was
shown that it can influence
classifier performance. We therefore JPEG compress all renders.

\noindent \textbf{Color Cast} For each channel of the image, with probability
$\frac{1}{2}$ we add a value sampled uniformly in $[-20, 20]$.

\noindent \textbf{Channel Swap} With probability $50\%$ we randomly swap the
color channels.

\noindent \textbf{Image degradation} Some ground truth bounding boxes are very
small. The resulting object resolution is very different
than our high resolution renders. In order for the model to learn to
classify correctly lower resolution images, we estimate the bounding
box area using the PASCAL training set, and~\ref{sub:render_quality}
down-sample $25\%$ of the
renders to a size that falls in the lower $30\%$ of the distribution.

\noindent \textbf{Occlusions} To get robustness to occlusions we
randomly place rectangular patches either from the PASCAL training set, or of
uniform color, on the renders. The size of the rectangle is
sampled between $0.2$ and $0.6$ of the render size.

\begin{figure}[t!]
  \centering
    \subfigure[Sample Images From RenderCar]{
      \includegraphics[width=0.40\linewidth]{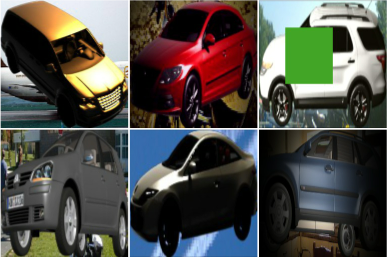}
      \label{FIG:renders_with_background}
   }
   \subfigure[Fully Rendered Scene]{
      \includegraphics[width=0.36\linewidth]{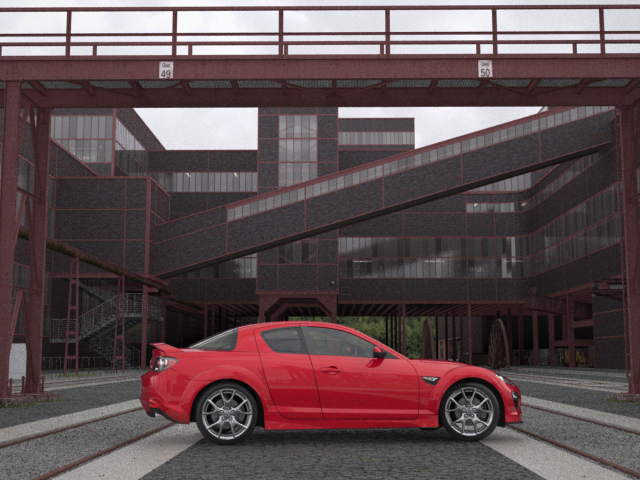}
      \label{FIG:render_scene_test}
   }
  \caption{
  (a) Rendered training images with data augmentation.
  (b) To evaluate the use of renders as test data we create
  scenes in which an object is placed in a fully modeled environment.
  This is challenging for
  models trained on natural images (Table~\ref{TABLE:dataset-comparisons}).
  }
\end{figure}

Finally, we perform a train/test split of the data such that images from 90
models are used for training, and one model is held out for testing. The
resulting datasets have 819,000 and 1,800 images respectively.
Figure~\ref{FIG:renders_with_background} shows a number of rendered images
after application of the data augmentation methods listed above. We name
this dataset RenderCar.

With the steady increase in computational power, the lowered cost of rendering
software, and the availability of 3D CAD models, for the first time it is now
becoming possible to fully model not just the object of interest, but also its
surrounding environment. Figure~\ref{FIG:render_scene_test} shows a fully
rendered images from one such scene. Rendering such a fully realistic image
takes considerably more time and computational resources than just the model.
Note the interaction between the model and the scene -- shadows, reflections,
etc. While we can not, at the moment, produce a large enough dataset of such
renders to be used for training purposes, we can utilize a smaller set of
rendered scene images for \emph{validation}. We create a dataset of fully rendered scenes which we term RenderScene. In
Table~\ref{TABLE:dataset-comparisons} we show that such a set is useful for
evaluating models trained on real images.

\section{Network Architecture and Loss Function}
\label{sec:model_renders4cv}

Following the success of deep learning based approaches in object detection and
classification, We base our network architecture on the widely used AlexNet
model~\cite{krizhevsky_imagenet_2012} with a number of modifications.
The most
important of those changes is our introduced loss function. Most previous work
on viewpoint estimation task have simply reduced it to a regular classification
problem. The continuous viewpoint space is discretized into a set of class
labels, and a classification loss, most commonly SoftMax, is used to train
the network. This approach has a number of appealing properties: (1) SoftMax is
a well understood loss and one that has been used successfully in many computer
vision tasks. (2) The predictions of a SoftMax are associated with probability
values that indicate the model's confidence in the prediction. (3) The
classification task is easier than a full regression to a real value angle which
can reduce over-fitting.

There is one glaring problem with this reduction - it does not take into account
the circular nature of angle values. The discretized angles are just treated as
class labels, and any mistake the model makes is penalized in the same way.
There is much information that is lost if the distance between the predicted
angle and the ground truth is not taken into account when computing the error
gradients. We use a generalization of the SoftMax loss
function that allows us to take into account distance-on-a-ring between the
angle class labels:

\begin{equation}
  E = -\frac{1}{N} \sum_{n=1}^N \sum_{k=1}^K w_{l_n, k} \log(p_{n, k}),
\end{equation}
where $N$ is the number of images, $K$ the number of classes, $l_n$ the ground
truth label of instance $n$, and $p_{n,k}$ the probability of the class $k$ in
example $n$. $w_{l_n, k}$ is a Von Mises kernel centered at $l_n$, with the width of the kernel controlled by $\sigma$:
\begin{equation}
\label{eq:von_mises}
  w_{l_n, k} = \exp(-\frac{\min (\abs{l_n-k}, K - \abs{l_n-k}) }{\sigma^2}).
\end{equation}

The Von Mises kernel implements a circular normal distribution centered at
$0\degree$. This formulation penalizes predictions that are far, in
angle space, more than smaller mistakes. By acknowledging the different types of
mistakes, more information flows back through the gradients. An
intuitive way to understand this loss, is as a matrix of class-to-class weights
where the values indicate the weights $w$. A
standard Softmax loss would have a weight of 1 on the diagonal of the matrix,
and 0 elsewhere. In the angle-aware version, some weight is given to
nearby classes, and there is a wrap-around such that weight
is also given when the distance between the predicted and true class crosses the
$0\degree$ boundary.

\section{Evaluation}
\label{sec:results_renders4cv}
Our objective is to evaluate the usefulness of rendered data
for training. First, we compare the result of the deep network
architecture described in Section~\ref{sec:model_renders4cv} on two fine-grained
viewpoint estimation datasets:

\noindent \textbf{CMU-Car}.
The MIT street scene data set~\cite{bileschi2006streetscenes} was augmented by
Boddeti et al.~\cite{boddeti2013correlation} with landmark annotations for
$3,433$ cars. To allow for evaluation of precise viewpoint estimation
Movshovitz-Attias et al.~\cite{movshovitzattias:2014:BMVC} further
augmented this data set by providing camera matrices for $3,240$ cars.
They manually annotated a 3D CAD car model
with the same landmark locations as the images and used the POSIT algorithm to
align the model to the images.

\noindent \textbf{PASCAL3D+}.
The dataset built by~\cite{xiang_beyond_2014} augments 12 rigid categories of the
PASCAL VOC 2012 with 3D annotations. Similar to above, the annotations were
collected by manually matching correspondence points from CAD models to images.
On top of those, more images are added for each category from the ImageNet
dataset. PASCAL3D+ images exhibit much more variability compared to the existing
3D datasets, and on average there are more than 3,000 object instances per
category. Most prior work however do not use the added ImageNet data and restrict
themselves to only the images from PASCAL VOC 2012.

We also define two synthetic datasets, which consist the the two types of
rendered images described in Section~\ref{sec:data_generation} -- \textbf{RenderCar} and
\textbf{RenderScene}. The RenderCar dataset includes images from the entire set of 3D CAD
models of vehicles. We use the various augmentation methods described above. For
RenderScene we use a single car model placed in a fully modeled environment
which depicts a fully realistic industrial scene as shown in
figure~\ref{FIG:render_scene_test}, rendered over all 1800 angles as described
in Section~\ref{sec:data_generation}.

We focus our evaluation on the process of viewpoint prediction, and work
directly on ground truth bounding boxes. Most recent work on detecting objects
and estimating their viewpoint, first employ an RCNN style bounding box
selection process~\cite{su2015render} so we feel our approach is reasonable.

We split each dataset into train/validation sets and train a separate deep model
on each one of the training sets. We then apply every model to the validation
sets and report the results. We perform viewpoint estimation on ground truth
bounding boxes for images of the car class. For PASCAL3D+ and CMUCar these
bounding boxes were obtained using annotators, and for the rendered images these
were automatically created as the tightest rectangle that contains all pixels
that belong to the rendered car.

\begin{figure}[t!]
  \centering
    \includegraphics[width=0.78\linewidth]{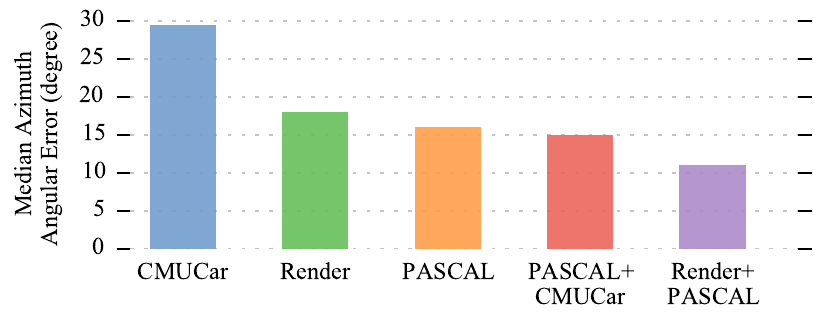}
    \caption{Median angular error on the PASCAL test set for
    a number of models. The model trained on PASCAL is the best model trained on
    a single dataset, but it has the advantage of having access to PASCAL images
    during training. Combining PASCAL images with rendered training data performs better then combining them with additional natural images from CMUCar.
    }
    \label{FIG:median_err_on_pascal}
\end{figure}

Figure~\ref{FIG:median_err_on_pascal} shows the median angular azimuth error of
4 models when evaluated on the PASCAL validation set. While the model trained on
PASCAL performs better than the one trained on rendered data it has an unfair
advantage - the rendered model needs to overcome domain adaptation as well as
the challenging task of viewpoint estimation. Note that the model trained on
rendered data performs much better than the one trained on CMUCar. To us this
indicates that some of the past concerns about generality of models trained from
synthetic data may be unfounded. It seems that the adaptation task from
synthetic data is not harder than from one real image set to another. Lastly
note that best performance is achieved when combining real data with synthetic
data. This model achieves better performance than when combining PASCAL data
with images from CMUCar. CMUCar images are all street scene images taken from
standing height. They have a strong bias, and add little to a model's
generalization.

\begin{figure*}[th]
  \centering
    \includegraphics[width=0.9\linewidth]{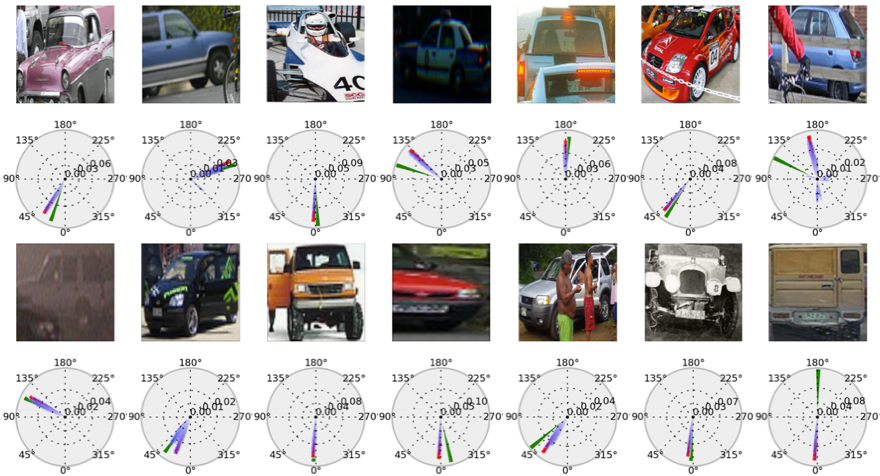}
    \vspace{0.005in}
    \caption{Sample results of our proposed method from
    the test set of PASCAL3D+. Below each image we show viewpoint
    probabilities assigned by the model (blue), predictions (red), and ground
    truth (green). Right column shows failure cases. Strongly directional occluders and
    $180\degree$ ambiguity are the most common failures.}
    \label{FIG:examples_viewpoint}
\end{figure*}

Figure~\ref{FIG:examples_viewpoint} shows example results on the
ground truth bounding boxes from the PASCAL3d+ test set. Successful predictions
are shown in the top two rows, and failure cases in the bottom row. The test set
is characterized by many cropped and occluded images, some of very poor image
quality. Mostly the model is robust to these, but when it mostly fails
due to these issues, or the $180\degree$ ambiguity.

\begin{table}[t!]
\centering
{\small
\scalebox{0.91}{
    \begin{tabular}{ c c c c c c | c c}
        & \multicolumn{6}{c}{\textbf{Validation}}\\
        \textbf{Training} & PASCAL & CMUCar & RenderCar & Render Full Scene& P+C & Avg & Avg On Natural\\
        \cmidrule(r){2-2} \cmidrule(r){3-3} \cmidrule(r){4-4} \cmidrule(r){5-5} \cmidrule(r){6-6} \cmidrule(l){7-7} \cmidrule(l){8-8}

        PASCAL (P) & 16\degree & 6\degree & 18\degree & 14\degree & 8\degree & 12.4\degree & 10\degree\\
        CMUCar (C)
        \colorcell& 29.5\degree \colorcell& \textbf{2\degree} \colorcell& 27\degree \colorcell& 13\degree \colorcell& 5\degree \colorcell& 15.3\degree \colorcell& 12.17\degree \colorcell\\
        RenderCar (R)
        & 18\degree& 6\degree& 2\degree& 8\degree& 8\degree & 8.4\degree& 10.67\degree\\

        P + C \colorcell& 15\degree \colorcell& 3\degree \colorcell& 13\degree \colorcell& 9\degree \colorcell& 5\degree \colorcell& 9\degree \colorcell& 7.67\degree \colorcell\\

        P + R& \textbf{11\degree}& 6\degree& 4\degree& 6\degree& 6\degree & 6.6\degree & 7.67\degree\\

        C + R \colorcell& 15\degree \colorcell& \textbf{2\degree} \colorcell& 2\degree \colorcell& \textbf{5\degree} \colorcell& 4\degree \colorcell& 5.6\degree \colorcell& 7\degree \colorcell\\

        P + C + R& 12\degree& \textbf{2\degree}& \textbf{1\degree}& 8\degree& \textbf{3\degree} & \textbf{5.2\degree} & \textbf{5.67}\degree\\

        \end{tabular}
}
}
\caption{Median angular error of car viewpoint estimation on ground truth bounding boxes. Note the distinct effect of dataset bias:
the best model on each dataset is the one trained on the corresponding training set.
On average, the model trained on
rendered images performs similarly to that trained on PASCAL, and better than
one that is trained on CMUCar. Combining rendered data with
natural images produces lower error than when combining two natural-image
datasets. Combining all three datasets provides lowest error. Last column shows
average error on columns 1,2,5.}

\label{TABLE:dataset-comparisons}
\end{table}

Table~\ref{TABLE:dataset-comparisons} shows model error for azimuth estimation
for all train/validation combinations. First, it is easy to spot dataset bias -
the best performing model on each dataset is the one that was trained on a
training set from that dataset. This is consistent with the findings
of~\cite{torralba2011unbiased} that show strong dataset bias
effects. It is also interesting to see how some datasets are better for
generalizing. The two right most columns average prediction error
across multiple datasets. The \emph{Avg} column averages across all datasets,
and \emph{Avg On Natural} averages the results on the 3 columns that use natural
images for testing. Notice that the model trained on PASCAL data performs better
overall than the one trained on CMUcar. Also notice that the model trained on
RenderCar performs almost as well as the one trained on PASCAL. When combining
data from multiple datasets the overall error is reduced. Unsurprisingly,
combining all datasets produces the best results. It is interesting to note,
however, that adding rendered data to one of the natural image datasets produces
better results than when combining the two real-image ones. We conclude that
this is because the rendered data adds two forms of variation: (1) It provides
access to regions of the label space that were not present in the small
natural-image datasets. (2) The image statistics of rendered images are different than
those of natural images and the learner is forced to generalize better in order
to classify both types of images. We further examine the effect of combining
real and synthetic data below.

\paragraph{Render Quality:} 
\label{sub:render_quality}
\begin{figure}[t!]
  \centering
    \includegraphics[width=0.87\linewidth]{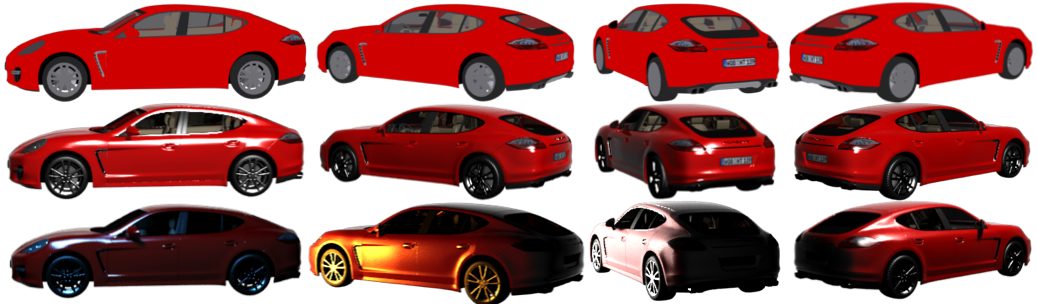}
    \caption{We evaluate the effect of render quality under 3
    conditions: simple object material, and ambient lighting
    (top row); complex material and ambient lighting
    (middle); and a sophisticated case using complex material, and
    directional lighting whose location, color, and strength are randomly
    selected (bottom).}
    \label{FIG:render_quality_examples}
\end{figure}

Renders can be created in varying degrees of quality and realism. Simple, almost
cartoon-like renders are fast to generate, while ones that realistically model
the interplay of lighting and material can be quite computationally expensive. Are these higher quality renders worth the added cost?

Figure~\ref{FIG:render_quality_examples} shows 3 render conditions we use to
evaluate the effect of render quality on system performance. The top row shows
the basic condition, renders created using simplified model materials, and
uniform ambient lighting. In the middle row are images created using a more
complex rendering procedure -- the materials used are complex. They more closely
resemble the metallic surface of real vehicles. We still use ambient lighting
when creating them. The bottom row shows images that were generated using
complex materials and directional lighting. The location, color, and strength of
the light source are randomly selected.

\begin{figure}[t!]
  \centering
    \includegraphics[width=0.73\linewidth]{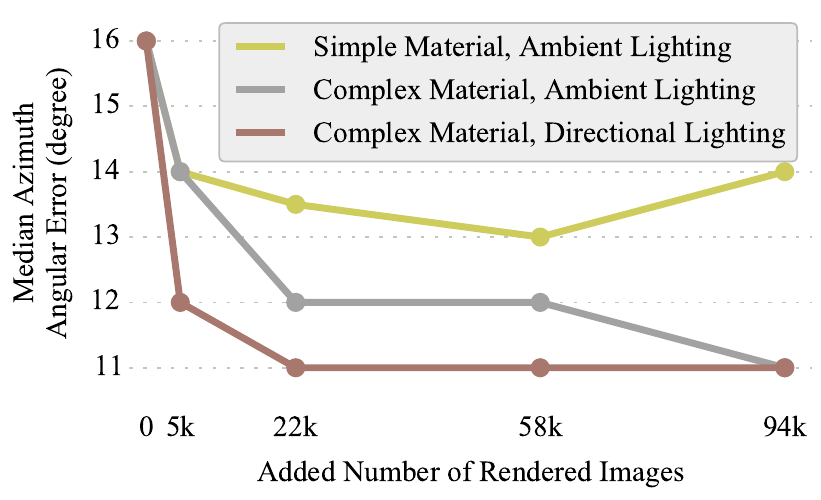}
    \caption{Error as a function of dataset size for the 3 render quality
    conditions described in Figure~\ref{FIG:render_quality_examples}. All
    methods add rendered images to the PASCAL training set. There is a decrease
    in error when complex model materials are used, and a further decrease when
    we add random lighting conditions.}
    \label{FIG:render_quality_err}
\end{figure}

Figure~\ref{FIG:render_quality_err} shows median angular error as a function of
dataset size for the 3 render quality conditions. We see that when the rendering
process becomes more sophisticated the error decreases. Interestingly, when
using low quality renders the error increases once the amount of renders
dominate the train set. We do not see this phenomena with higher
quality renders. We conclude that using complex materials and lighting is
an important aspect of synthetic datasets.

\paragraph{Balancing a Training Set Using Renders:} 
\label{sub:balancing}
\begin{figure*}[ht]
  \centering
    \includegraphics[width=1.0\linewidth]{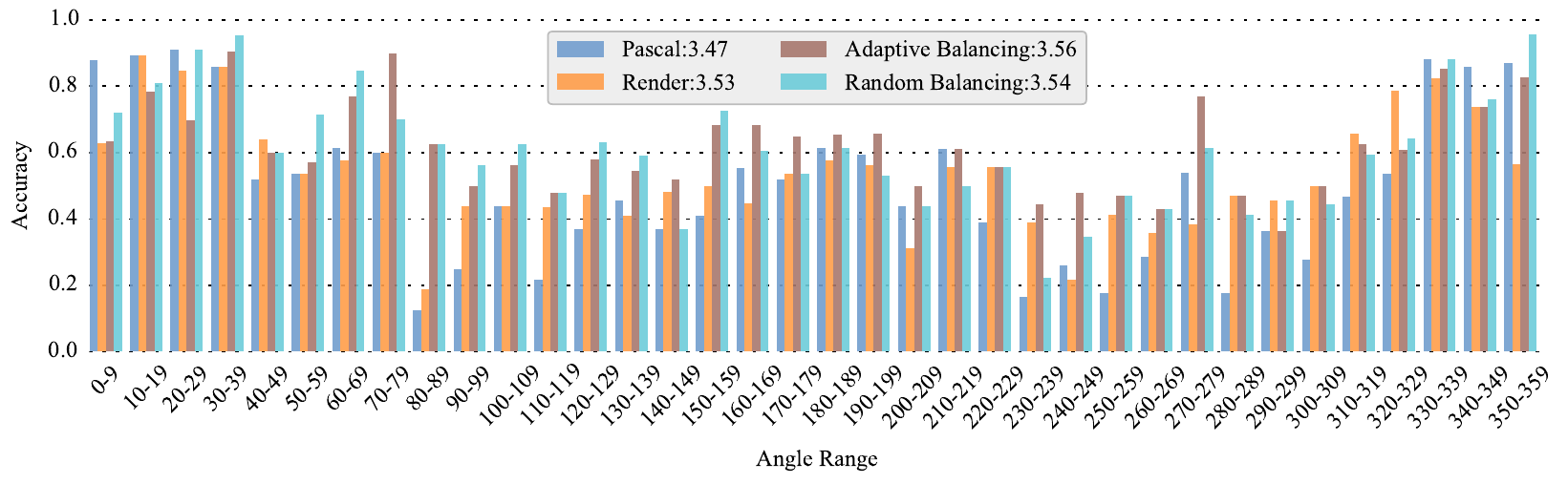}
    \caption{
    Model accuracy as a function of ground truth angle.
    A model should perform uniformly well across all angles.
    The entropy of each distribution (legend) shows deviation from uniform.
    For the model trained on PASCAL images (blue),
    the accuracy mirrors the train set distribution
    (Figure~\ref{FIG:angle_freq_compare}).
    The entropy of a model trained on rendered images is higher, but still not uniform (orange). This is
    due to biases in the test set.
    Brown and light blue bars show accuracy distribution of models for which rendered
    images were used to balance the training set. We see they have higher entropy
    with adaptive balancing having the highest. Models were tested on PASCAL test images.
    }
    \label{FIG:adaptive_balancing}
\end{figure*}
So far we have seen that combining real images with synthetically generated ones
improves performance. Intuitively this makes sense, the model is given more
data, and thus learns a better representation. But is there
something more than that going on? Consider a trained model. What are the
properties we would want it to have? We would naturally want it to be as
accurate as possible. We would also want it to be unbiased -- to be
accurate in all locations in feature/label space. But bias in the training set
makes this goal hard to achieve. Renders can be
useful for bias reduction.

\begin{table}[t]
{\small
\centering
    \begin{tabular}{ @{}C{2.5cm} C{1.5cm} C{1.5cm} C{3cm} C{3cm}}
        \textbf{Training Set:} & PASCAL & Render & Adaptive Balancing & Random Balancing \\
         \textbf{MAE:} & $26.5\degree$ & $26.0\degree$ & $\textbf{16.0\degree}$ & $17.5\degree$ \\

        \end{tabular}
}
\vspace{0.035in}
\caption{Median angular error (MAE) on a viewpoint-uniform sample of the PASCAL
test set. Performance with rendered based training set is best.}
\label{TABLE:balanced_test_set}
\end{table}

In Figure~\ref{FIG:adaptive_balancing} we quantify this requirement. It shows
the accuracy of 3 models as a function of the angle range in which
they are applied. For example, a model trained on 5k PASCAL images (top row),
has about 0.8 accuracy when it is applied on test images
with a ground truth label in $[0,9]$.

We want the shape of the accuracy distribution to be as close to uniform as
possible. Instead, we see that the model performs much better on some angles,
such as $[0,30]$, than others, $[80,100]$. If the shape of the distribution
seems familiar it is because it closely mirrors the training set distribution
shown in Figure~\ref{FIG:angle_freq_compare}. We calculate the models' entropy on
the accuracy distribution entropy as a tool for comparison. Higher entropy indicates
a more uniform distribution. When
using a fully balanced training set of rendered images (2nd row), the test
entropy increases. This is encouraging, as the model contains less angle
bias.

Rendered images can be used as a way to balance the training set to reduce bias
while still getting the benefits of natural images. We experiment with two
methods for balancing which we call adaptive balancing (3rd row), and random
balancing (bottom row). In adaptive balancing we sample each angle reversely
proportional to its frequency in the training set. This method transforms the
training set to uniform using the least number of images. In random balancing
the added synthetic images are sampled uniformly. As the ratio of rendered
images in the training set increases, the set becomes more uniform.
For this experiment both methods added 2,000 images to the PASCAL training set.
Interestingly, the two balancing methods have a similar prediction entropy.
However, no method's distribution is completely uniform. From that we can
conclude that there are biases in the test data other than angle distribution,
or that some angles are just naturally harder to predict than others.

These results raise an interesting question: how much of the performance gap we
have seen between models trained on real PASCAL images, and those trained on
rendered data stems from the angle bias of the test set?
Table~\ref{TABLE:balanced_test_set} shows the median errors of these models on a
sample of the PASCAL test set in which all angles are equally represented. Once
the test set is uniform we see that models based on a balanced training set
perform best, and the model trained solely on PASCAL images has the worst
accuracy.

Lastly, we hold constant the number of images used for training and vary the
proportion of rendered images. Table~\ref{TABLE:blend_ratio} shows the error of
models trained with varying proportions of real-to-rendered data. Models are
trained using 5,000 images -- the size of the PASCAL training set. Notice that
there is an improvement when replacing up to 50\% of the images with rendered
data. This is likely due to the balancing effect of the renders on the angle
distribution which reduces bias. When most of the data is rendered, performance
drops. This is likely because the image variability in renders is smaller than
real images. More synthetic data is needed for models based purely on it to
achieve the lowest error -- 18\degree.

\paragraph{Size of Training Set:} 
\label{sub:size_of_training_set}
We examine the role of raw number of renders in performance. We
keep the number of CAD models constant at 90, and uniformly sample renders
from our pool of renders. Many of the cars in PASCAL are only
partially visible, and we want the models to learn this, so in each set half the
rendered images show full cars, and half contain random $60\%$ crops
Table~\ref{TABLE:training_set_size} summarizes the results of this experiment.
Better performance is achieved when increasing the number of renders, but there
are diminishing returns. There is a limit to the variability a model can learn
from a fixed set of CAD models, and one would need to increase the size of the
model set to overcome this. Obtaining a large amount of 3D models can be
expensive and so this motivates creation of large open source model datasets.

\begin{table}[t!]
\centering
{\small
    \begin{tabular}{ C{4cm} C{0.5cm} C{0.67cm} C{0.67cm} C{0.67cm} C{0.8cm} }
         \textbf{Renders in Train Set:}& $0\%$ & $25\%$ & $50\%$ & $75\%$ & $100\%$ \tabularnewline
         \textbf{Median Angular Error:} & $16\degree$ & $\textbf{14\degree}$ & $\textbf{14\degree}$ & $15\degree$ & $22\degree$ \tabularnewline
    \end{tabular}
}
\vspace{0.035in}
\caption{Median angular error (MAE) on PASCAL test set. Training set size
is fixed at 5k images (the size of the PASCAL training set) and we modify the
proportion of renders used. There is an improvement when replacing up to
$\frac{1}{2}$ of the images with renders. The renders help balance the angle
distribution and reduce bias.}
\label{TABLE:blend_ratio}
\end{table}

\begin{table}[t!]
\centering
\scalebox{0.9}{
    \begin{tabular}{@{}C{2.5cm} C{0.5cm} C{1cm} C{0.75cm} C{0.8cm} C{0.5cm} C{1cm} C{0.5cm} C{0.8cm}}
        Train Set Size & \multicolumn{2}{c}{10k} & \multicolumn{2}{c}{50k} & \multicolumn{2}{c}{330k} & \multicolumn{2}{c}{820k}\\
        Loss Layer& SM & wSM & SM & wSM & SM & wSM & SM & wSM\\
        \cmidrule(r){1-1} \cmidrule(r){2-3} \cmidrule(r){4-5} \cmidrule(r){6-7} \cmidrule{8-9}
        MAE & $31\degree$ & $27.5\degree$ & $54.5\degree$ & \hspace{0.15cm}$21\degree$ & $26\degree$ & $21.5\degree$ & $23\degree$ & $\textbf{18\degree}$\\
        \end{tabular}
}
\caption{Effect of training set size and loss layer. We see a trend of
better performance with increased size of training set, but the effect quickly
diminishes. Clearly, more than just training set size
influences performance. The Von Mises weighted SoftMax (wSM) performs better
than regular SoftMax (SM) for all train sizes.}
\label{TABLE:training_set_size}
\end{table}

\paragraph{Loss Layer:} 
\label{sub:loss_layer}
When using Von Mises kernel-based SoftMax loss
function is that it is impossible to obtain zero loss. In a regular SoftMax, if
the model assigns probability 1.0 to the correct class it can achieve a loss
of zero. In the weighted case, when there is some weight assigned to more
than one class, a perfect prediction will actually result in infinite loss,
as all other classes have probability
values of zero, and the log-loss assigned to them will be infinite. Minimum loss
will be reached when prediction probabilities are spread out across nearby
classes. This is both a downside of this loss function, but also its strength.
It does not let the model make wild guesses at the correct class. Nearby
views are required to have similar probabilities. It trades
angle resolution with added prediction stability.

Table~\ref{TABLE:training_set_size} compares the results of SoftMax based
models and models trained using the weighted SoftMax layer over a number of
rendered dataset sizes. The weighted loss layer performs better for all
sizes. This supports our hypothesis that viewpoint estimation is not a standard classification task.
Most experiments in this section were performed using $\sigma = 2$ for the Von
Mises kernel in Equation~\eqref{eq:von_mises}.
This amounts to an effective width of $6\degree$, meaning that predictions that
are farther away from the ground truth will not be assigned any
weight. Table~\ref{TABLE:sigma} shows the effect of varying this value.
Interestingly we see that the method is robust to this parameter, and
performs well for a wide range of values. It appears that even a relatively weak
correlation between angles provides benefits.

\begin{table}[t!]
\begin{minipage}[c]{0.48\textwidth}
\centering
\scalebox{0.9}{
    \begin{tabular}{c c c c c c}
        $\sigma $& 2 & 3 & 4 & 10 & 15 \\
        Effective Width & $6\degree$ & $8\degree$ & $12\degree$ & $30\degree$ & $50\degree$ \\
        \cmidrule{2-6}
        Median Angular Error & 13\degree & 14\degree & 14\degree & \textbf{13\degree} & 15\degree\\
        \end{tabular}
}
\caption{Effect of kernel width ($\sigma$) of the Von Mises kernel. The loss
function is not sensitive to the selection of kernel width.}
\label{TABLE:sigma}
\end{minipage}
\hfill
\begin{minipage}[c]{0.48\textwidth}
\centering
\scalebox{0.9}{
    \begin{tabular}{c c c c c c c}
        \% Occluded Images& 0.0 & 0.1 & 0.35 & 0.4 & 0.5 & 1.0 \\
        \cmidrule{2-7}
        MAE & 25\degree & 25\degree & \textbf{21\degree} & 25\degree & 28\degree & 26\degree\\
        \end{tabular}
}
\caption{Effect of added occlusions to the training set. All models evaluated
used datasets of 50,000 rendered images.}
\label{TABLE:occlusions}
\end{minipage}
\end{table}

\paragraph{Occlusion:} 
\label{sub:occlusion}
The PASCAL test set contains many instances of partially occluded cars. When
augmenting synthetic data we add randomly sized occlusions to a subset of
the rendered images. Table~\ref{TABLE:occlusions} shows that there is some benefit from
generating occlusions, but when too many of the training images are occluded it
becomes harder for the model to learn. What is the best way to generate
occlusions? In our work we have experimented with simple, single color,
rectangular occlusions, as well as occlusions based on random patches from
PASCAL. We saw no difference in model performance. It would be interesting to
examine the optimal spatial occlusion relationship. This is likely to be object
class dependent.

\section{Discussion} 
\label{sec:render4cv_discussion}
In this work we propose the use of rendered images
as a way to automatically build datasets for viewpoint estimation.
We show that models trained on renders are competitive with
those trained on natural images -- the gap in performance can be explained by
domain adaptation. Moreover, models
trained on a combination of synthetic/real data outperform ones trained on
natural images.

The need for large scale datasets is growing with the increase in
model size. Based on the results detailed here we believe that
synthetic data should be an important part of dataset creation strategies. We
feel that a combination a small set of carefully annotated images,
combined with a larger number of synthetic renders, with automatically assigned
labels, has the best cost-to-benefit ratio.

This strategy is not limited to viewpoint estimation, and can be employed for a
range of computer vision tasks. Specifically we feel that future research should
focus on human pose estimation, depth prediction, wide-baseline correspondence
learning, and structure from motion.


\clearpage

\bibliographystyle{splncs03}
\bibliography{render4cv}
\end{document}